\documentclass[letterpaper,times]{IONconf}
\usepackage[pdftex]{graphicx}
\usepackage{url}

\usepackage[
  backend=biber,
  style=apa,
  maxcitenames=2,  
  mincitenames=1,
  maxbibnames=99,  
  uniquelist=false
]{biblatex}
\usepackage[hidelinks]{hyperref}

\usepackage{mathtools}
\usepackage{amssymb}
\usepackage{bm}
\usepackage{bbm}
\usepackage{caption}
\usepackage{subcaption}
\usepackage{xcolor}
\usepackage{authblk}
\usepackage{siunitx}
\usepackage{booktabs}
\usepackage{multirow}
\usepackage{algorithm}
\usepackage{algorithmic}
\usepackage{float} 

\title{Full Stack Navigation, Mapping, and Planning for the Lunar Autonomy Challenge}

\author{Adam~Dai, Asta~Wu, Keidai~Iiyama, Guillem~Casadesus~Vila,  Kaila~M.~Y.~Coimbra, Thomas~Deng, and Grace~Gao}
\affil{\textit{Stanford University}}

\begin{document}
\newcommand{\todo}[1]{{\color{Red}TODO: #1}}

\newcommand{\red}[1]{{\color{red}{#1}}}

\newcommand{\R}{\ensuremath{\mathbb{R}}}

\newcommand{\SO}{\mathbf{SO}}
\newcommand{\SE}{\mathbf{SE}}
\newcommand{\so}{\mathfrak{so}}
\newcommand{\se}{\mathfrak{se}}

\maketitle

\section*{Abstract}

We present a modular, full-stack autonomy system for lunar surface navigation and mapping developed for the Lunar Autonomy Challenge. Operating in a GNSS-denied, visually challenging environment, our pipeline integrates semantic segmentation, stereo visual odometry, pose graph SLAM with loop closures, and layered planning and control. We leverage lightweight learning-based perception models for real-time segmentation and feature tracking and use a factor-graph backend to maintain globally consistent localization. High-level waypoint planning is designed to promote mapping coverage while encouraging frequent loop closures, and local motion planning uses arc sampling with geometric obstacle checks for efficient, reactive control. We evaluate our approach in the competition's high-fidelity lunar simulator, demonstrating centimeter-level localization accuracy, high-fidelity map generation, and strong repeatability across random seeds and rock distributions. Our solution achieved first place in the final competition evaluation.

\section{Introduction}
\label{sec:intro}

As renewed interest in space exploration accelerates, there is a growing need for autonomous robotic systems capable of operating in harsh, remote environments with limited human oversight. Onboard autonomy plays a critical role in enabling navigation, mapping, and scientific decision-making in environments where communication delays or outages preclude constant teleoperation. NASA’s Curiosity and Perseverance Mars rovers have already demonstrated the value of autonomy for surface missions, and upcoming lunar programs—such as the Endurance rover~\parencite{baker2024endurance} and the CADRE multi-robot system~\parencite{de2024multi}—seek to further advance these capabilities. Autonomy not only improves efficiency and safety for individual agents, but also enables scalable multi-robot operations, supporting the broader goals of mission campaigns, such as the NASA Artemis program.

The lunar environment poses unique challenges that make robust autonomous navigation especially difficult. In contrast to structured terrestrial environments, lunar rovers must traverse unstructured, previously unseen terrain without access to GPS or pre-built maps. Hazards such as sharp rocks, loose regolith, and steep slopes can threaten safe traversal. Perception is further complicated by extreme lighting: the absence of an atmosphere produces high-contrast shadows with sharp edges and overexposed regions that degrade the performance of vision-based systems. In many regions, the surface is textureless and feature-sparse, making localization and mapping particularly challenging.

In this paper, we present a full-stack autonomous agent for lunar rover navigation and mapping, developed for the Lunar Autonomy Challenge~\parencite{LunarAutonomyChallenge}. The Lunar Autonomy Challenge is a collaboration between NASA, The Johns Hopkins University (JHU) Applied Physics Laboratory (APL), Caterpillar Inc., and Embodied AI. The challenge is managed by APL for NASA. The challenge supports the Lunar Surface Innovation Initiative (LSII) by advancing technologies required for sustained surface operations. The competition uses a high-fidelity simulator built with Unreal Engine and CARLA~\parencite{CARLA}, featuring realistic vehicle dynamics and photorealistic imagery of lunar terrain. Teams are tasked with mapping a 27~m $\times$ 27~m region around a simulated lander using an autonomous digital twin of NASA’s ISRU Pilot Exacavator (IPEx) rover.

Our solution integrates semantic perception, stereo visual odometry, pose graph SLAM with loop closures, and hierarchical planning into a modular pipeline. We evaluate the system in diverse simulated environments and demonstrate strong mapping and localization performance under variable lighting, terrain, and initial conditions. This work culminated in a first-place finish in the final competition evaluation.

\subsection*{Key Contributions}

\begin{enumerate}
    \item We develop a modular lunar rover autonomy stack integrating semantic segmentation, stereo visual odometry, pose graph optimization, and hierarchical planning.
    \item We propose a structured path planning strategy that promotes coverage and loop closure for consistent mapping.
    \item We demonstrate robust localization and mapping performance across varying terrain conditions, lighting settings, and random seeds.
    \item We open-source our implementation to support reproducibility and further development by the community.\footnote{Code: \url{https://github.com/Stanford-NavLab/lunar_autonomy_challenge}}
\end{enumerate}
\section{Lunar Autonomy Challenge}
\label{sec:background}

The Lunar Autonomy Challenge is a NASA and JHU APL–led competition designed to advance autonomy for surface missions. Teams deploy a digital twin of NASA’s IPEx rover in a high-fidelity Unreal Engine simulator and must autonomously map a 27~\si{m}~$\times$~27~\si{m} region around a lander, estimating both terrain elevation and rock locations on a discrete grid. Scoring emphasizes accurate geometric mapping within 5 cm tolerance and reliable rock detection, under harsh lighting and feature-sparse lunar conditions.
All information and images below are taken from the official documentation at \texttt{\url{https://lunar-autonomy-challenge.jhuapl.edu/Challenge-Documentation}}; please refer to it for full details.

\begin{figure}[ht]
    \centering
    \begin{minipage}{0.47\textwidth}
        \centering
        \includegraphics[width=\linewidth]{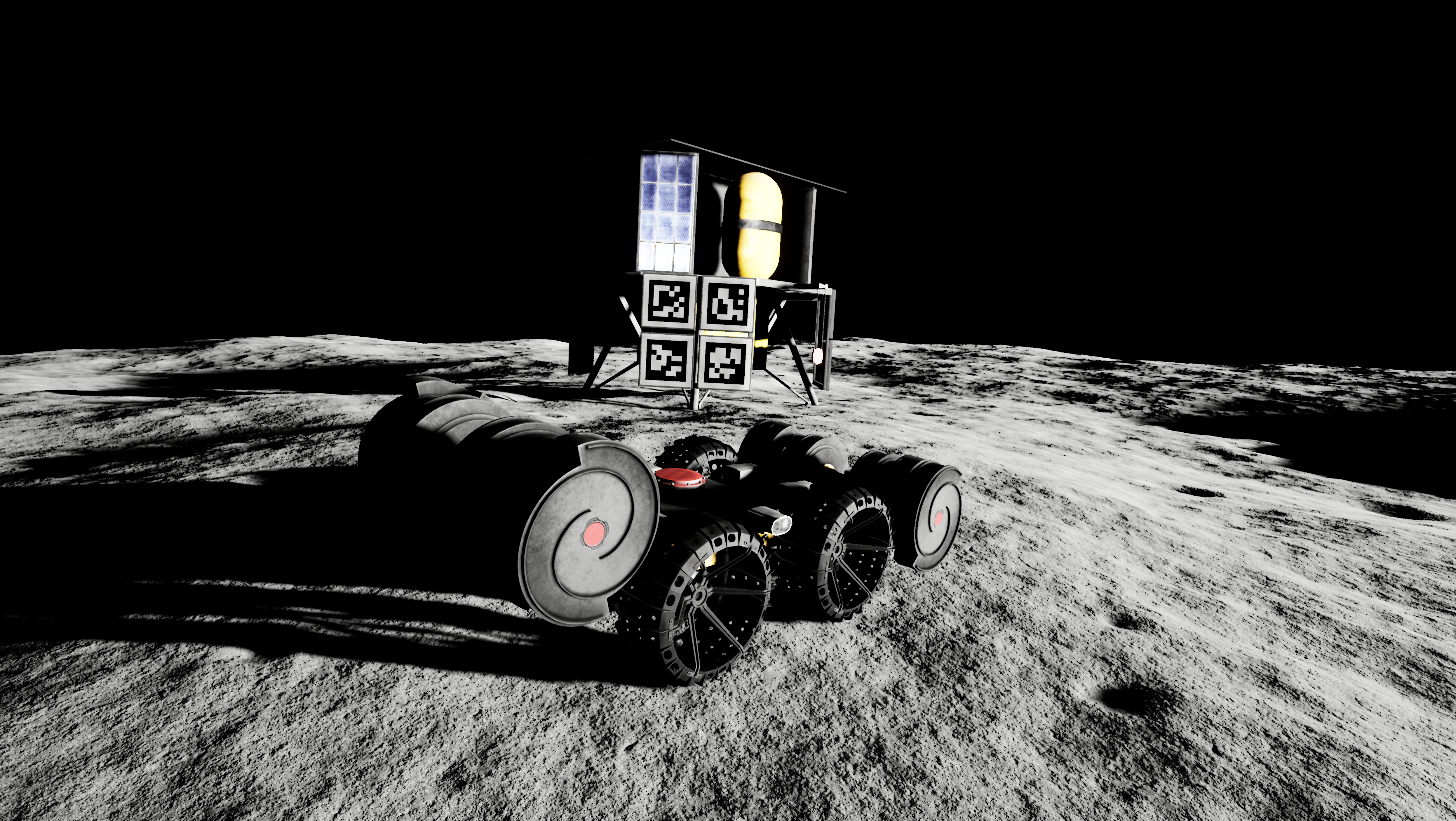} 
        \subcaption{View from the simulator showing the lunar robot with the lander in the background.}
        \label{fig:LAC_robot}
    \end{minipage}\hfill
    \begin{minipage}{0.5\textwidth}
        \centering
        \includegraphics[width=\linewidth]{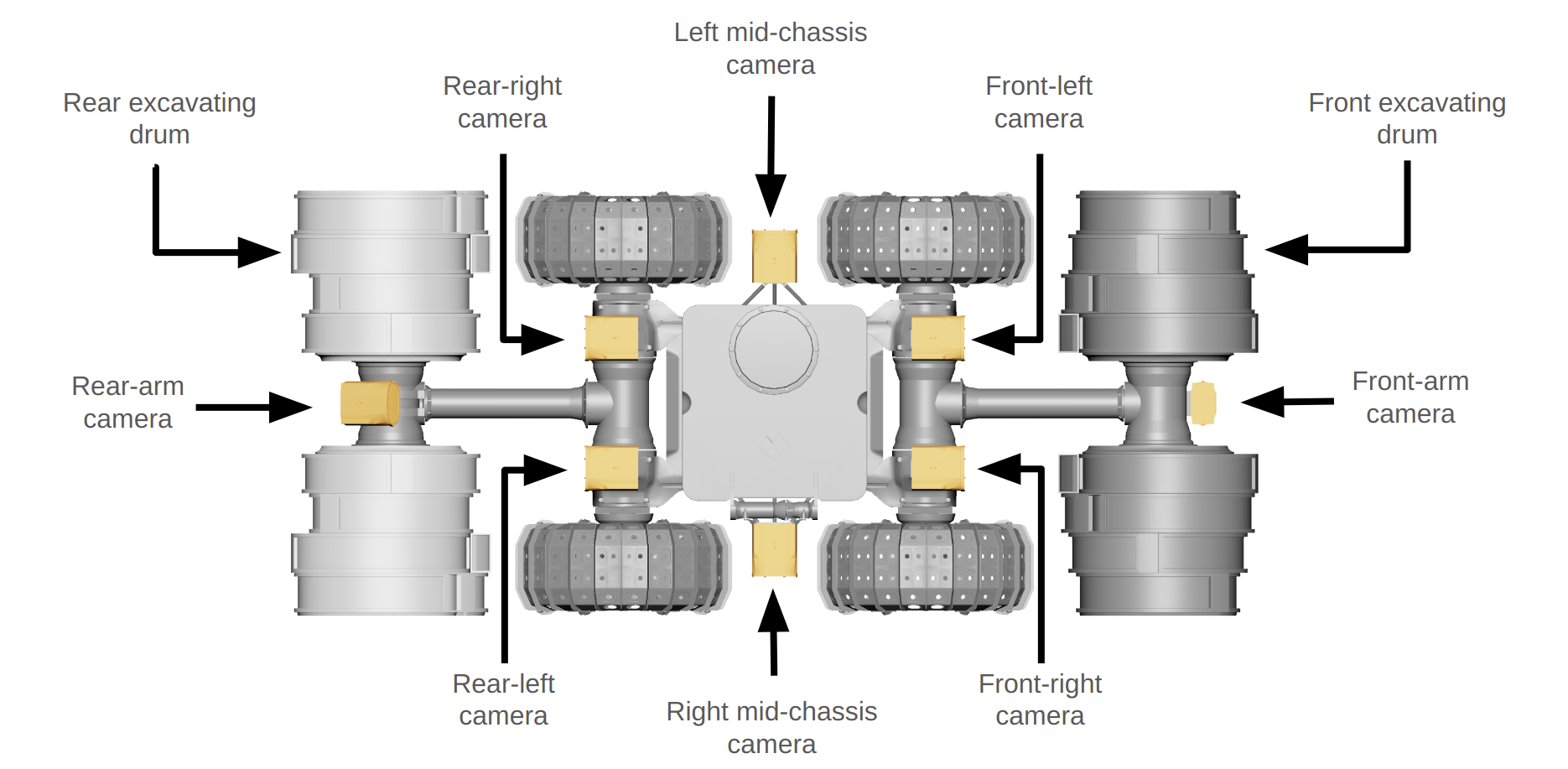}  
        \subcaption{Diagram of the lunar robot showing the arrangement of its eight cameras.}
        \label{fig:camera_geometry}
    \end{minipage}
    \begin{minipage}{0.8\textwidth}
        \centering
        \includegraphics[width=\linewidth]{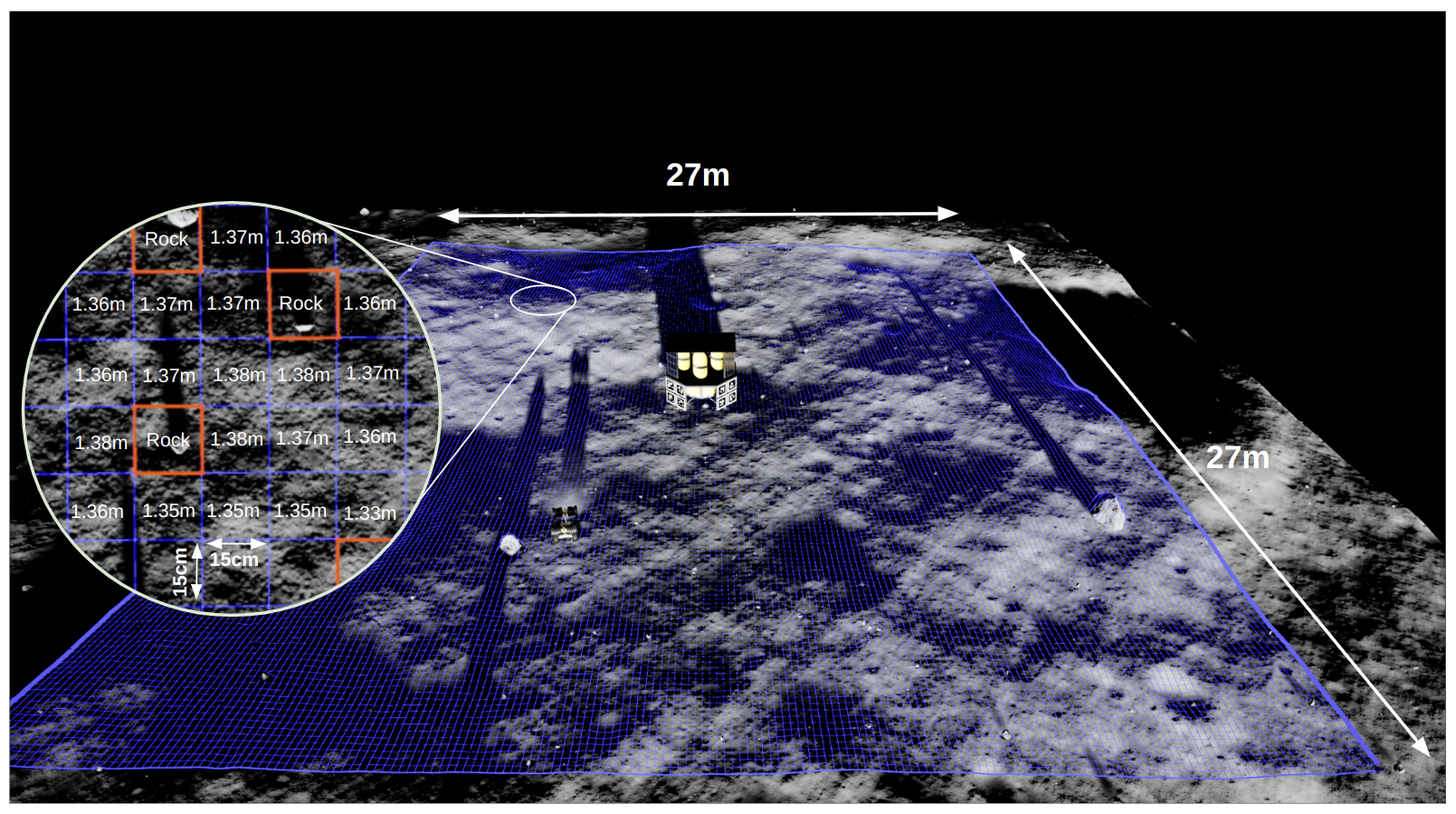}  
        \subcaption{Mapping area for the challenge. The objective is to map both terrain height as well as rock presence within a 27~\si{m}~$\times$~27~\si{m} area around the lunar lander.}
        \label{fig:LAC_mapping_area}
    \end{minipage}
    \caption{Key aspects of the Lunar Autonomy Challenge: (a) realistic image rendering, (b) realistic rover design with actuation, and (c) competition mapping objective. Images are adapted from the Lunar Autonomy Challenge simulator and documentation~\parencite{LunarAutonomyChallenge}}
    \label{fig:LAC_overview}
\end{figure}

\subsection{Lunar Robot}
The lunar robot is a four-wheeled robot with differential steering modeled after NASA's IPEx robot and equipped with an Inertial Measurement Unit (IMU) and eight monochrome cameras for perception.
The cameras are arranged in front and back stereo pairs, left and right side-facing cameras, and one mounted on each front and back arm, as shown in Figure~\ref{fig:camera_geometry}.
Each camera is coupled with an LED light in the same enclosure which can be used to illuminate the surroundings.
The robot is also equipped with articulating arms that support rotating drums at the front and rear for regolith excavation, but the 2025 version of the Lunar Autonomy Challenge does not involve excavation and focuses on mapping only.

\subsection{Simulator and Environment}

The simulator runs at 20 Hz and generates IMU data and ground truth robot pose (to be used for development and testing; disabled for the competition) at each frame.
It is capable of rendering images at 10 Hz from all 8 cameras of the IPEx robot at user-specified resolution, with a maximum resolution of 2448 $\times$ 2048 pixels.
In addition, the simulator allows for rendering ground-truth semantic masks in development mode.
The user may also configure the activation and intensity of the robot's LED lights.

The simulation environment consists of a 40 \si{m} $\times$ 40 \si{m} region of lunar terrain, with a lunar lander positioned at the center of the map.
The lander has a charging station for re-charging the robot as well as fiducial markers which may be used to assist the robot's localization.
The simulator provides approximate initial poses of both the robot and lander at initialization.
Figure~\ref{fig:LAC_robot} shows the lunar robot together with the lander as rendered by the simulator.

\subsection{Objectives and Scoring}

The objective of the challenge is to map a 27 \si{m} $\times$ 27 \si{m} region centered around the lander, shown in Figure~\ref{fig:LAC_mapping_area}.
The map is represented as a 180 $\times$ 180 discrete grid of cells, with cell resolution of 15 \si{cm} $\times$ 15 \si{cm}, and consists of a geometric height component and a rock component.
For each cell, the geometric map stores the estimated average height, while the rock map stores a boolean value indicating rock presence.
The geometric map is scored based on the number of cell heights correctly estimated to within a 50 \si{mm} (0.05 \si{m}) tolerance, whereas the rock is scored based on the F1 score in equation~\eqref{eq:f1rock}
\begin{equation}
    S_{rock} = \frac{2 \times TP}{2 \times TP + FP + FN} \label{eq:f1rock}
\end{equation}
where $TP$ is the number of true positive cells, $FP$ is the number of false positive cells, and $FN$ is the number of false negative cells.
Additionally, points are awarded for disabling the lander fiducials, and for completing the mapping within the allotted mission time of 24 hours.
\section{Related Work}
\label{sec:related_work}

\subsection{Full Autonomy Stacks in Challenging Environments}

Large-scale autonomy challenges such as the DARPA Subterranean (SubT) Challenge have demonstrated that fully autonomous robotic systems can operate in complex, GPS-denied environments. 
Team CERBERUS~\parencite{tranzatto2022cerberus} fielded a heterogeneous team of quadrupeds, aerial robots, and rovers, integrating multi-modal perception (LiDAR, RGB, thermal, IMU), multi-robot mapping, and robust planning to achieve state-of-the-art performance in underground exploration. 
Similarly, Team CoSTAR’s NeBula~\parencite{agha2021nebula} architecture developed a modular autonomy stack for heterogeneous teams, emphasizing resilient navigation, mapping, and decision-making under perceptual uncertainty. 
These efforts highlight that full autonomy stacks are feasible in highly unstructured environments, but they typically rely on rich multi-sensor payloads and multi-robot coordination. 
In contrast, the Lunar Autonomy Challenge constrains agents to vision and IMU only, while further stressing perception with extreme lighting and feature-sparse terrain. 
Our work proposes a novel autonomy stack tailored to these lunar-specific conditions.

\subsection{Visual SLAM}

Visual SLAM has a long history in robotics, with methods such as ORB-SLAM~\parencite{campos2021orbslam3} establishing reliable feature-based monocular and stereo pipelines for real-time localization and mapping. Extensions like Kimera~\parencite{rosinol2020kimera} demonstrate the value of tightly integrating metric-semantic reconstruction and trajectory estimation for richer 3D scene understanding. More recently, learning-based approaches such as DROID-SLAM~\parencite{teed2021droid} and successors have achieved state-of-the-art accuracy through deep neural architectures for dense correspondence and motion estimation. However, most benchmarks for these systems—such as KITTI, EuRoC, and TUM RGB-D—feature structured, textured environments that provide abundant features for localization. Under extreme lighting or texture-poor conditions, these methods still degrade or fail, limiting their applicability to planetary environments.

\subsection{Learned Feature Matching}

Recent progress in learned keypoint detectors and matchers has introduced methods such as SuperPoint~\parencite{detone_superpoint_2018-1} and LightGlue~\parencite{lindenberger2023lightglue}, which improve robustness to viewpoint and illumination changes. While still emerging in SLAM pipelines, these components represent a promising direction for addressing feature scarcity and perceptual challenges. Our system explicitly incorporates such learned features into the visual odometry and SLAM front-end, demonstrating their utility in the harsh lunar domain.
\section{Approach}
\label{sec:approach}


\begin{figure}[ht]
    \centering
    \includegraphics[width=\linewidth]{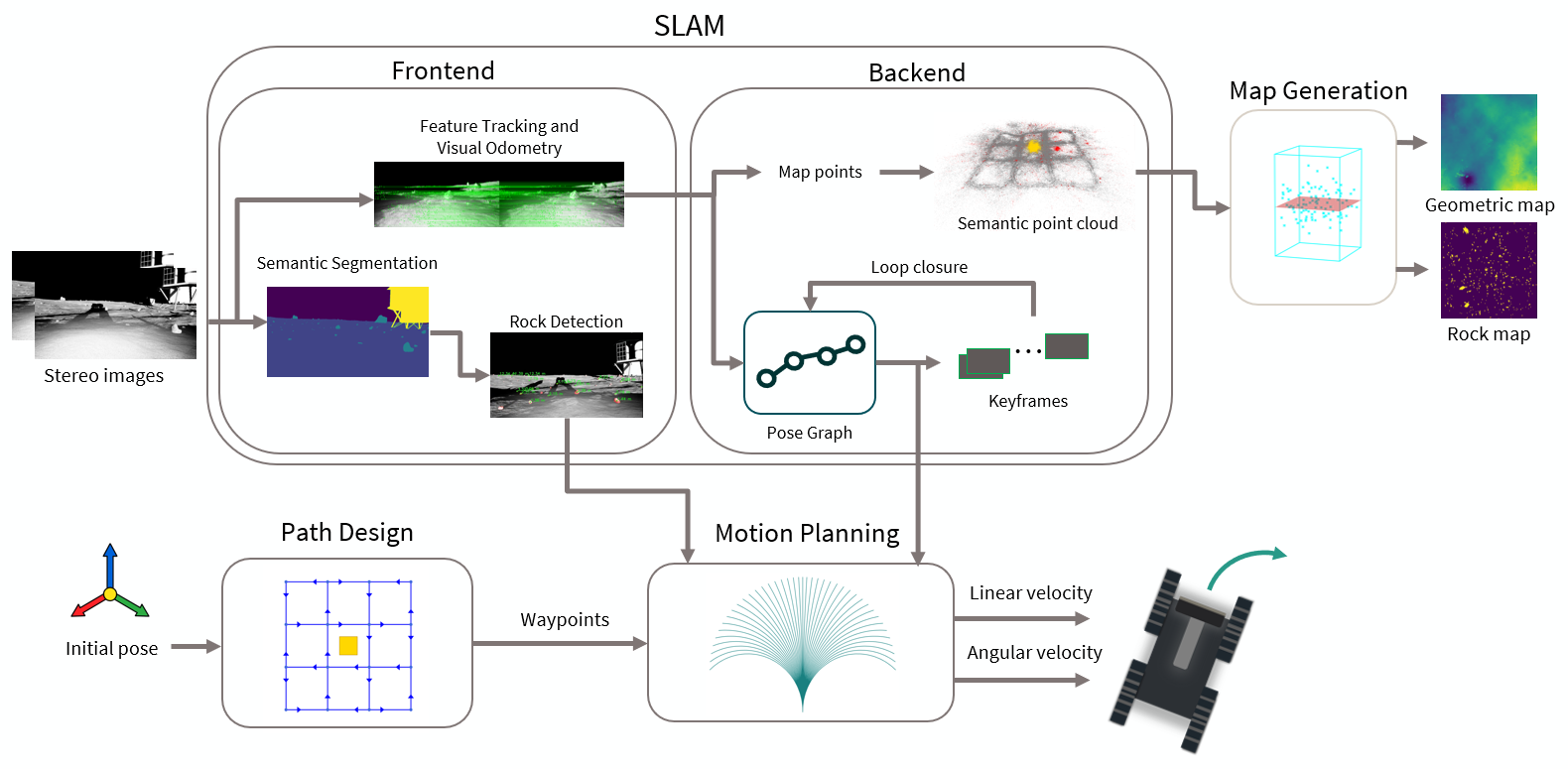}  
    \caption{
    Block diagram of our overall approach. 
    Stereo images are processed by the front-end for semantic segmentation, rock detection, and stereo visual odometry. 
    Detected rocks are passed to the motion planner, while tracked 3D features and odometry estimates are used in the SLAM backend for pose graph construction and loop closure.
    Optimized poses and semantic landmarks are projected into a global frame to generate geometric and rock maps. 
    High-level path design generates waypoints that encourage loop closures and full map coverage, while the motion planner selects safe arcs using geometric rock avoidance.
    }
    \label{fig:block_diagram}
\end{figure}

Our high-level approach, illustrated in Figure~\ref{fig:block_diagram}, is a modular and layered autonomy stack. 
Stereo images are processed by a front-end perception module that performs semantic segmentation, detects rocks, and extracts features for stereo visual odometry. 
These observations are passed to the SLAM backend, which constructs a pose graph and applies loop closures to maintain globally consistent localization. 
Simultaneously, detected rocks and the current estimated pose are used by the motion planner to select safe trajectories from sampled arcs. 
A high-level planner generates goal waypoints in a structured loop pattern to promote mapping coverage and loop closure. 
The final output of the system includes semantic maps of terrain geometry and rock presence, which are produced by projecting labeled 3D landmarks into a global grid. 
In the following sections we will describe each of the components in more detail.


\subsection{Semantic Segmentation}
\label{subsec:segmentation}
We use semantic segmentation to classify each image pixel into relevant terrain categories, including ground, rock, lander, fiducials, and sky. Segmentation plays a key role in both mapping and navigation: rock detection is essential for generating the binary rock occupancy map required by the challenge, and also enables obstacle avoidance during path planning by identifying hazardous terrain. We selected U-Net++~\parencite{zhou_unet_2018} as our segmentation model due to its strong performance across semantic segmentation benchmarks and efficient inference characteristics. To tailor it to the lunar domain, we fine-tuned the model on semantic masks generated by the simulator, using a dataset of approximately 5000 labeled images with standard augmentations. Segmentation is performed on the front-facing stereo images at 10 Hz, and the resulting masks are used to filter feature points, detect obstacles, and support downstream mapping.

\begin{figure}[ht]
    \centering
    \includegraphics[width=\linewidth]{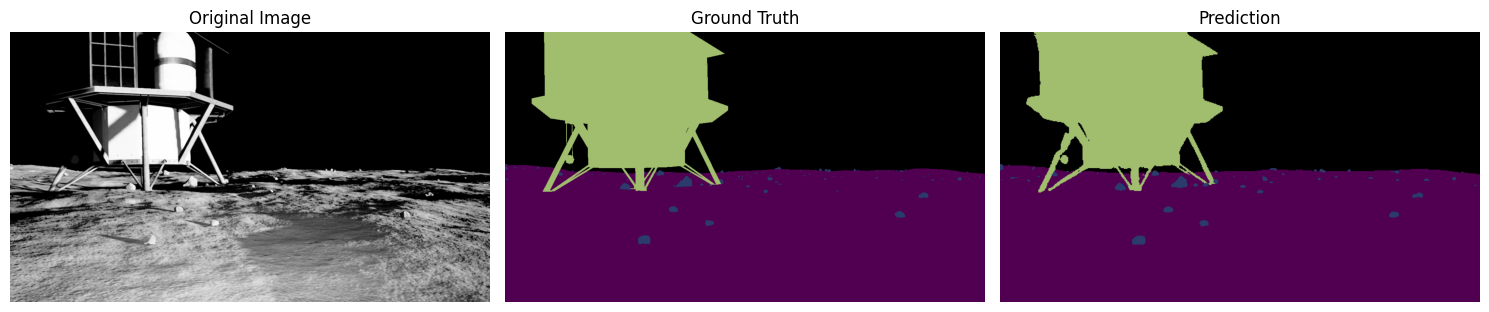}
    \caption{Semantic segmentation example. Given input image, each pixel is labeled with a semantic class of sky, ground, rock, lander, or fiducials.}
    \label{fig:segmentation}
\end{figure}


\subsection{Feature Extraction and Matching}
\label{subsec:features}
\begin{figure}[ht]
    \centering
    \includegraphics[width=0.5\linewidth]{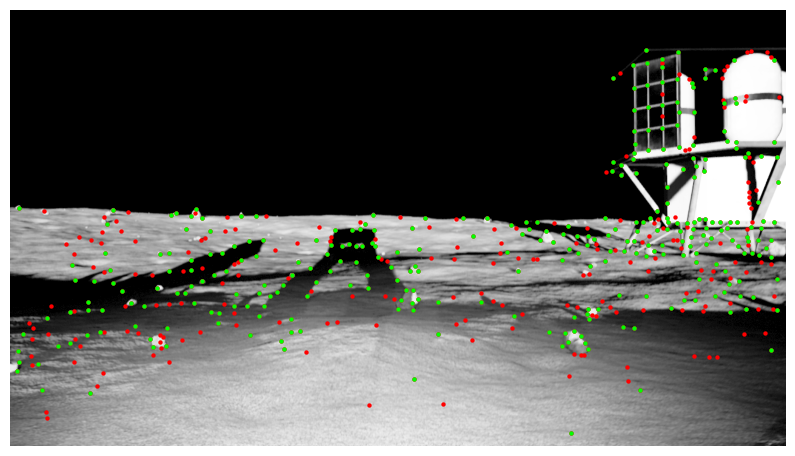}
    \caption{Extracted keypoints. Points in green have a score greater than 0.5, while points in red have score less than 0.5.} 
    \label{fig:keypoints}
\end{figure}

We use SuperPoint~\parencite{detone_superpoint_2018-1} for keypoint detection and description. SuperPoint is a self-supervised convolutional neural network that operates over the entire image to jointly predict 2D keypoints and their associated descriptors. Unlike traditional detectors that use hand-crafted features (e.g., SIFT, ORB), SuperPoint is highly efficient and robust to changes in lighting and viewpoint, making it well-suited for the lunar domain. Given an input image, SuperPoint returns a set of 2D keypoints in pixel coordinates, along with per-keypoint descriptors and detection confidence scores. An example of extracted keypoints is shown in Figure~\ref{fig:keypoints}.

\begin{figure}[ht]
    \centering
    \includegraphics[width=0.9\linewidth]{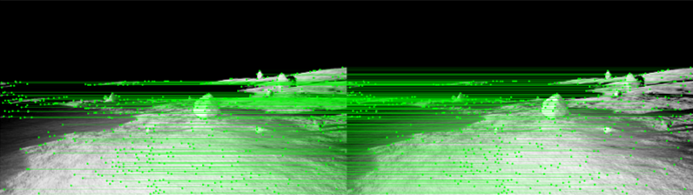}
    \caption{LightGlue feature matching between left and right stereo images.} 
    \label{fig:feat_matching}
\end{figure}

For matching, we use LightGlue~\parencite{lindenberger2023lightglue}, a transformer-based architecture designed for learned feature matching. LightGlue takes two sets of keypoints and their descriptors as input and outputs correspondences by jointly reasoning about geometric consistency and feature similarity. It is designed to be robust under extreme changes in illumination, scale, and perspective, which is particularly important given the harsh lighting conditions of the lunar surface. The matched keypoint pairs are used to establish geometric constraints between views.
An example of feature matching is shown in Figure~\ref{fig:feat_matching}.

\begin{figure}[ht]
    \centering
    \includegraphics[width=0.9\linewidth]{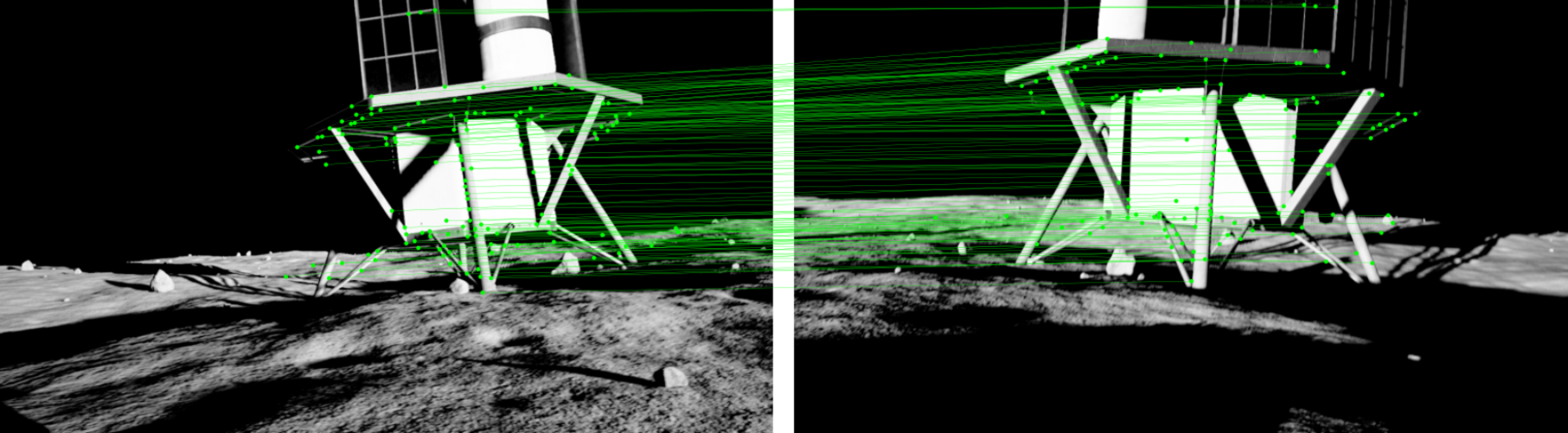}
    \caption{LightGlue feature matching on the lander between different viewpoints.} 
    \label{fig:feat_matching_lander}
\end{figure}

Feature extraction and matching serve as the backbone of our visual odometry pipeline and are critical for stereo triangulation, frame-to-frame tracking, and loop closure detection within our SLAM system.



\subsection{Stereo Visual Odometry and Tracking}
\label{subsec:vo}

We implement stereo visual odometry (VO) by triangulating 3D landmarks from stereo image pairs and tracking them across time for motion estimation. Given a rectified stereo pair, we extract features using SuperPoint and match them using LightGlue to obtain dense stereo correspondences. Using known camera intrinsics and baseline, we compute disparity for each matched keypoint and back-project them into 3D in the rover frame. These 3D points, along with their associated 2D keypoints and descriptors, form the input to our tracking and pose estimation pipeline.

Our VO and feature tracking system is implemented as a persistent stateful module. 
Specifically, we maintain a set of tracked features over time which includes unique track IDs, 2D keypoints, triangulated 3D coordinates, semantic labels, descriptors, and the number of frames each point has been successfully tracked. 
This persistent state enables reliable 2D–3D correspondences for estimating frame-to-frame motion using Perspective-n-Point (PnP). 
PnP is solved using OpenCV's~\parencite{bradski2000opencv} \texttt{solvePnPRansac} function.

\begin{figure}[ht]
    \centering
    \includegraphics[width=0.9\linewidth]{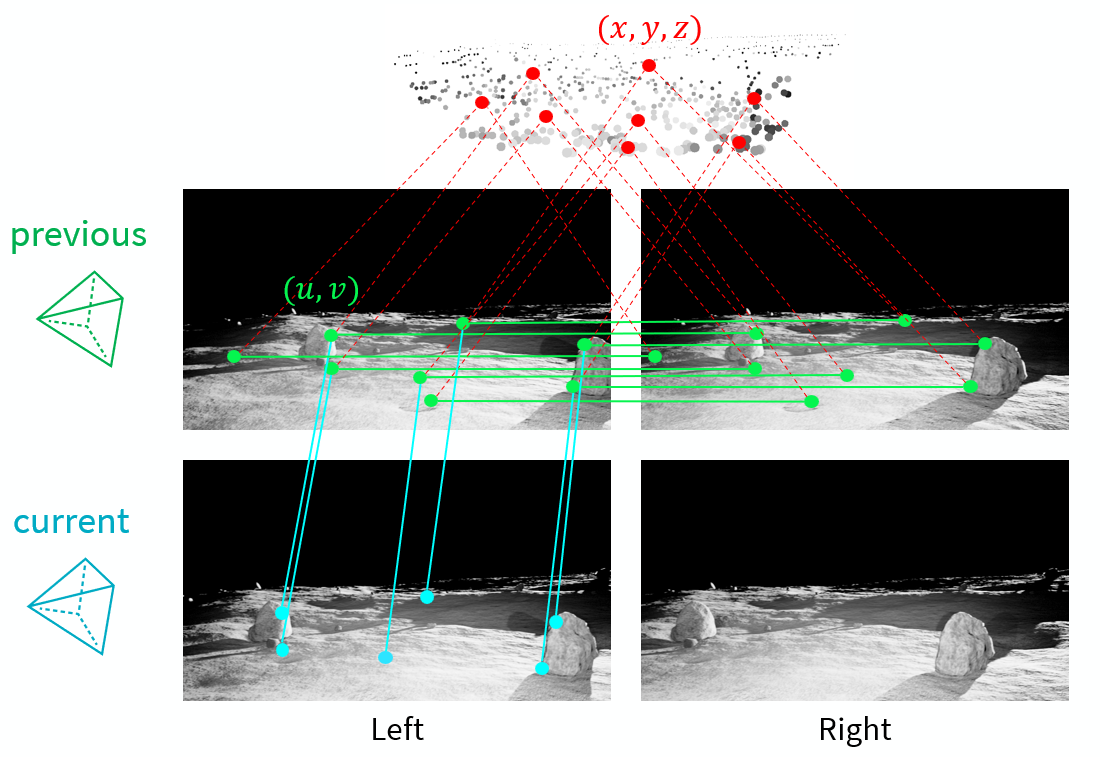}
    \caption{Visualization of the feature matching used to perform visual odometry. Stereo matched features are extracted from left and right frames, and the matched left image features in the previous frame are matched with that of the current frame. The PnP problem of triangulated 3D points in previous frame and matched 2D points in current frame is solved to obtain relative pose of the camera from previous to current frame.} 
    \label{fig:vo}
\end{figure}

At each frame, we perform the following steps:
\begin{enumerate}
    \item \textbf{Stereo Initialization:} When the tracker is first initialized, we match features between the left and right stereo images to compute depth and initialize 3D landmarks. Semantic labels are assigned to each landmark based on the segmentation mask.
    \item \textbf{Feature Matching:} For subsequent frames, features from the current left image are matched to the previous frame using descriptor similarity. Matches are used to associate current 2D keypoints with previously triangulated 3D points. This process is depicted in Figure~\ref{fig:vo}.
    \item \textbf{Pose Estimation:} Given the resulting 2D–3D correspondences, we solve a PnP problem to estimate the relative camera pose between frames.
    \item \textbf{Track Management:} Tracked points are updated based on match results. Existing tracks are extended when matched, and new tracks are initialized for unmatched keypoints. 3D points and semantic labels are updated using the latest observations.
\end{enumerate}

This tightly coupled VO and tracking pipeline serves as the core of our localization front-end, providing accurate relative pose estimates at 10 Hz. It also supports loop closure detection and map construction by maintaining consistent and semantically labeled 3D feature tracks over time.

\subsection{Simultaneous Localization and Mapping (SLAM)}
\label{subsec:slam}

Accurate localization was critical for achieving high-quality maps in the challenge. We implement a feature-based SLAM system that fuses stereo visual odometry measurements with periodic loop closures to maintain a globally consistent trajectory. Our SLAM architecture is modular and divided into a \textit{frontend} and \textit{backend}, where the frontend processes sensor data to produce relative motion estimates and semantic observations, and the backend performs pose graph optimization over selected keyframes.

\subsubsection{Frontend}

The SLAM frontend aggregates outputs from previous modules, including semantic segmentation and visual odometry. It processes images from the front stereo pair at 10 Hz and inertial measurements at 20 Hz. The main outputs of the frontend are:
\begin{itemize}
    \item Odometry estimates $\mathbf{T}_{k-1 \rightarrow k} \in \text{SE}(3)$ between consecutive frames via stereo VO and PnP.
    \item A set of tracked 3D feature points with semantic labels.
    \item Detected rocks used for updating the occupancy map.
\end{itemize}

These outputs are logged and passed to the backend to construct and optimize the global pose graph.

\begin{figure}[ht]
    \centering
    \includegraphics[width=\linewidth]{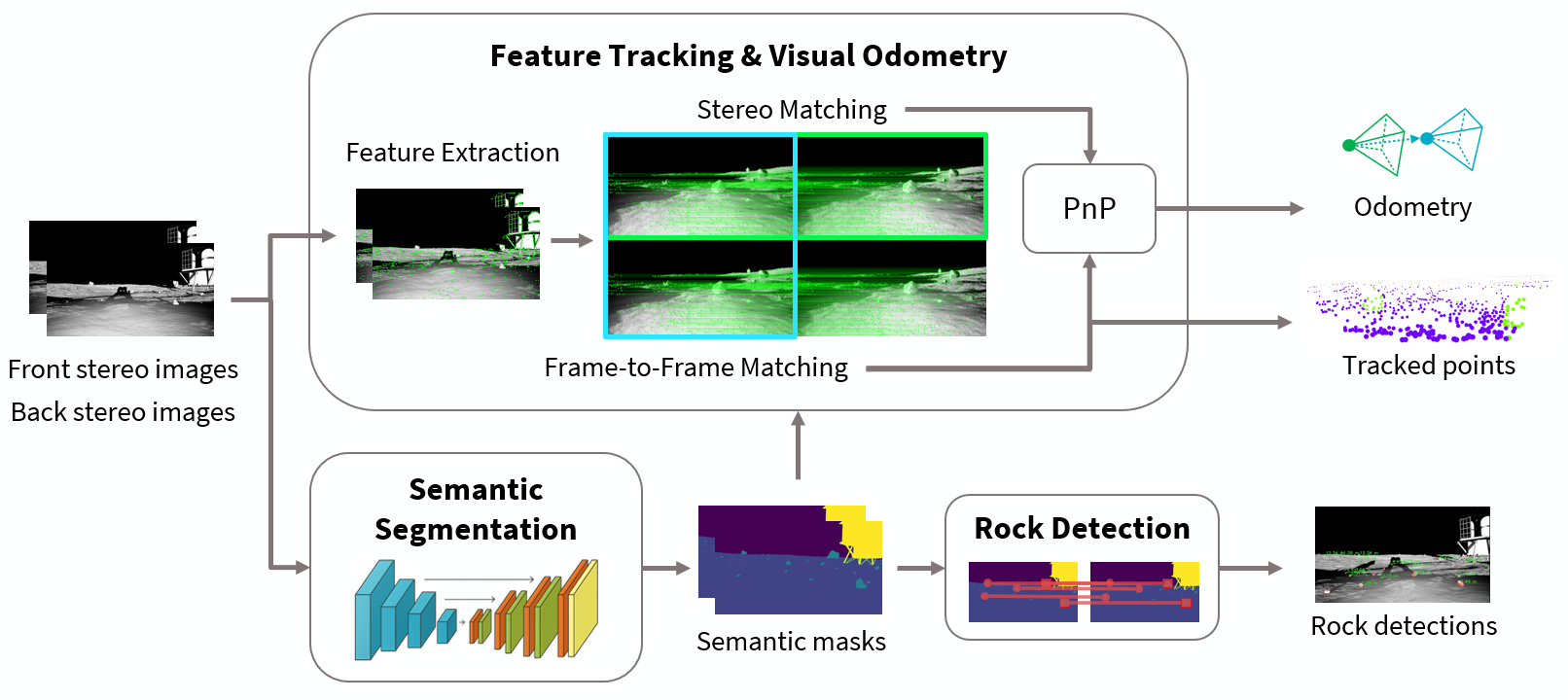}
    \caption{Overview of the SLAM frontend. Front and back stereo images and fed through feature tracking and visual odometry to produce an odometry estimate and tracked points. Semantic segmentation is used to extract semantic masks used to label tracked points and detect rocks for collision avoidance.} 
    \label{fig:frontend}
\end{figure}

\subsubsection{Backend}

The backend maintains a factor graph composed of camera poses $\mathbf{T}_k \in \text{SE}(3)$ and relative motion constraints from visual odometry or IMU. For each new frame, a pose is initialized by composing the previous pose with the odometry estimate:
\[
\mathbf{T}_k = \mathbf{T}_{k-1} \cdot \Delta \mathbf{T}_{k-1 \rightarrow k}
\]
A corresponding odometry factor is added to the graph:
\[
\mathbf{z}_{k-1,k}^{\text{odom}} = \Delta \mathbf{T}_{k-1 \rightarrow k}, \quad \Sigma_{\text{odom}}
\]
The odometry noise model $\Sigma_{\text{odom}}$ is chosen based on the source (VO or IMU).

In addition to pose updates, we associate each pose with a set of newly initialized 3D feature points $\{ \mathbf{x}_i \}$ in the rover frame, triangulated from stereo matches and semantically labeled. These are stored per-pose and later transformed into world coordinates using the optimized pose:
\[
\mathbf{x}_i^{(w)} = \mathbf{T}_k \cdot \mathbf{x}_i
\]

Every 20th frame is designated as a keyframe. For each keyframe, we store stereo features and descriptors for potential loop closure detection. We exclude the most recent $N$ keyframes and evaluate older keyframes based on:
\begin{itemize}
    \item \textbf{Translation}: Euclidean distance between positions must be below a threshold:
      \[
      \| \mathbf{t}_{k_i} - \mathbf{t}_{k_j} \|_2 < \tau_t
      \]
    \item \textbf{Rotation}: Angular difference (based on rotation matrix error) must satisfy:
      \[
      \| \log( \mathbf{R}_{k_j}^\top \mathbf{R}_{k_i} ) \|_2 < \tau_r
      \]
\end{itemize}

If a candidate passes both checks, we attempt to compute a relative pose using stereo PnP between stored keyframe features and the current image. If successful, we add a loop closure factor:
\[
\mathbf{z}_{k_j,k_i}^{\text{loop}} = \mathbf{T}_{k_j \rightarrow k_i}, \quad \Sigma_{\text{loop}}
\]
The factor graph is then re-optimized using Levenberg–Marquardt with GTSAM~\parencite{gtsam}. This ensures long-term consistency and is crucial for correcting drift.

Finally, we project all semantically labeled local 3D landmarks from the graph into world coordinates, producing a global semantic point cloud:
\[
\mathcal{P}_\text{map} = \bigcup_k \mathbf{T}_k \cdot \mathcal{P}_k
\]
This point cloud is passed to the mapping module for conversion into rock and height maps. The latest pose $\mathbf{T}_k$ and estimated velocity (via finite differencing of recent poses) are provided to the planner.

\begin{algorithm}[H]
\caption{SLAM Backend Update}
\label{alg:slam_backend}
\begin{algorithmic}[1]
\REQUIRE Odometry estimate $\Delta \mathbf{T}$, tracked points $\mathcal{P}$, left/right images, semantic mask
\STATE Insert new pose $\mathbf{T}_k = \mathbf{T}_{k-1} \cdot \Delta \mathbf{T}$ into graph
\STATE Add odometry factor between $X_{k-1}$ and $X_k$
\STATE Add newly triangulated semantic points $\mathcal{P}_k$ to point map
\IF{frame is keyframe}
    \STATE Save stereo features and descriptors
    \STATE Detect loop closure candidates based on distance and angle
    \FOR{each loop closure candidate}
        \STATE Attempt stereo PnP to estimate $\mathbf{T}_{\text{loop}}$
        \IF{successful}
            \STATE Add loop closure factor
        \ENDIF
    \ENDFOR
    \STATE Optimize pose graph using Levenberg–Marquardt
\ENDIF
\end{algorithmic}
\end{algorithm}


\subsection{Geometric and Rock Map Generation}
\label{subsec:map_generation}

The final output of our SLAM and perception pipeline is a dense semantic point cloud in the world frame, where each point is associated with an $(x, y, z)$ position and a semantic class label (e.g., ground or rock). From this point cloud, we generate two required maps: a geometric elevation map and a binary rock occupancy map. Both maps are represented as $180 \times 180$ grids centered around the lander, with a cell resolution of $15~\text{cm} \times 15~\text{cm}$.

\subsubsection{Geometric Map}  
To compute the geometric elevation map, we filter the semantic point cloud to retain only points classified as \textit{ground}. Each ground point is projected to its corresponding 2D grid cell based on its $(x, y)$ position. For each cell, we collect the set of $z$-values (elevations) of all points that fall into it and compute the median:
\[
z_{i,j} = \text{median} \left( \{ z_k \mid (x_k, y_k, z_k) \in \text{Ground}, \, (i,j) = \pi(x_k, y_k) \} \right)
\]
where $\pi(\cdot)$ denotes the projection from world coordinates to grid cell indices. The use of the median operator makes the elevation estimate more robust to outliers and noise in the point cloud.

\subsubsection{Rock Map}  
To compute the binary rock occupancy map, we use a probabilistic majority-vote strategy. For each grid cell, we count the number of points labeled as \textit{rock} and \textit{ground}. Let $n_\text{rock}$ and $n_\text{ground}$ be the number of points of each class in a given cell. The final label $l_{i,j}$ is assigned as:
\[
l_{i,j} = 
\begin{cases}
1 & \text{if } n_\text{rock} > n_\text{ground} \\
0 & \text{otherwise}
\end{cases}
\]
This simple thresholding rule corresponds to a maximum a posteriori (MAP) estimate under a Dirichlet-Categorical model with uniform prior, where each observed point contributes to a count over the discrete rock/ground classes. This formulation provides a principled probabilistic interpretation while remaining efficient and interpretable.


\subsection{High-Level Planning}
\label{subsec:path_design}

We use a fixed high-level waypoint generation strategy designed to maximize mapping coverage while encouraging frequent loop closures. The resulting waypoints are passed to the local motion planner, which executes short-horizon collision-avoiding trajectories.

Our path design assumes only the rover's initial position and operates in a fixed frame centered around the lander. The trajectory begins with a small loop around the lander to initialize tracking and accumulate early keyframes. It then progresses outward by tracing the perimeters of nested $3 \times 3$ grid cells, with each loop covering one square in the grid. The initial loop covers the central cell, and subsequent loops expand outward (e.g., left-middle, top-middle, right-middle, etc.).

Each new loop is designed to:
\begin{itemize}
    \item \textbf{Overlap with adjacent loops} — ensuring feature re-observations.
    \item \textbf{Revisit previous areas} — encouraging reliable loop closures.
    \item \textbf{Maintain visual contact with the lander} early on — improving initialization stability.
\end{itemize}

This strategy produces dense coverage of the required $27 \times 27$ meter mapping region with minimal drift accumulation, since the rover frequently returns to previously mapped regions from multiple directions.

\begin{figure}[ht]
    \centering
    \begin{subfigure}[t]{0.19\textwidth}
        \includegraphics[width=\linewidth]{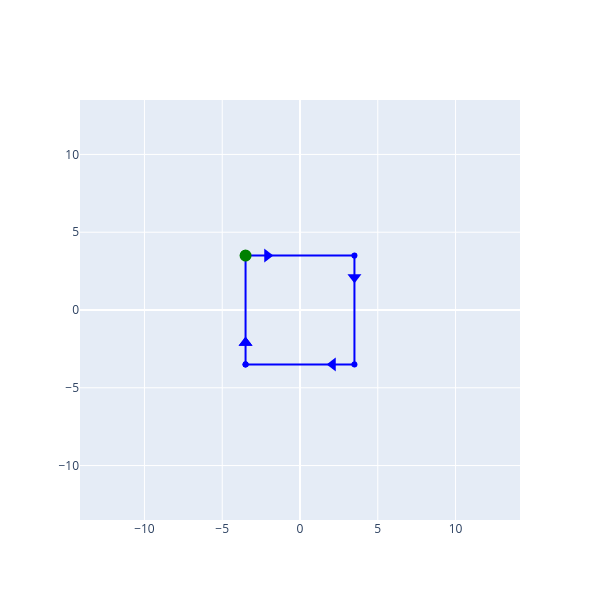}
        \caption{Initial loop}
    \end{subfigure}
    \hfill
    \begin{subfigure}[t]{0.19\textwidth}
        \includegraphics[width=\linewidth]{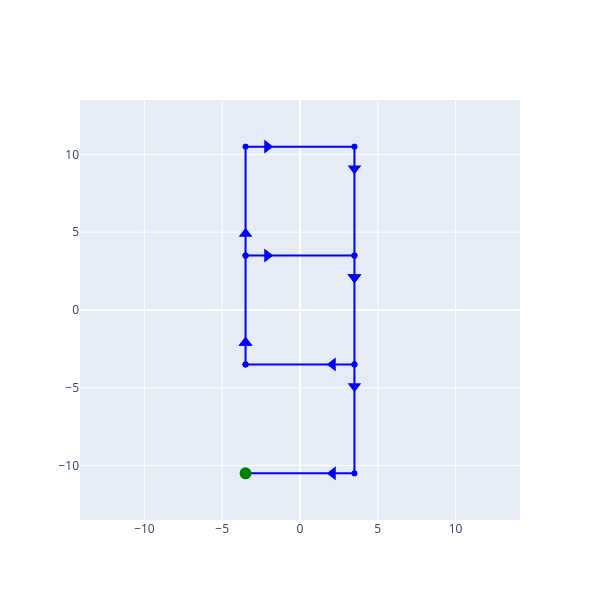}
        \caption{Expanding outward}
    \end{subfigure}
    \hfill
    \begin{subfigure}[t]{0.19\textwidth}
        \includegraphics[width=\linewidth]{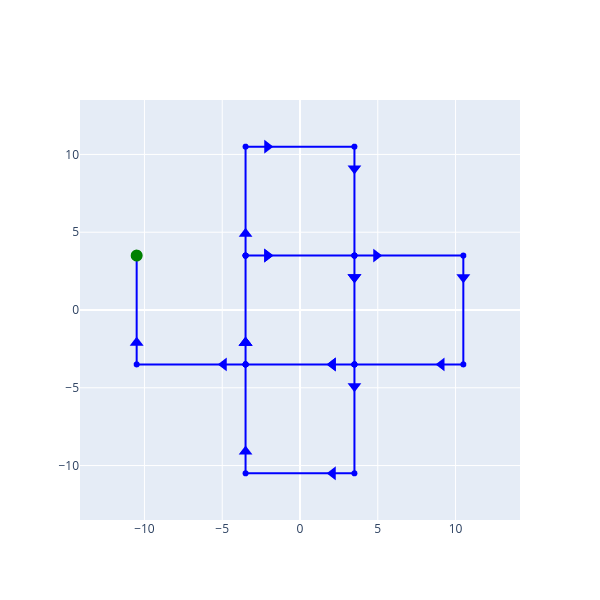}
        \caption{Cross segments}
    \end{subfigure}
    \hfill
    \begin{subfigure}[t]{0.19\textwidth}
        \includegraphics[width=\linewidth]{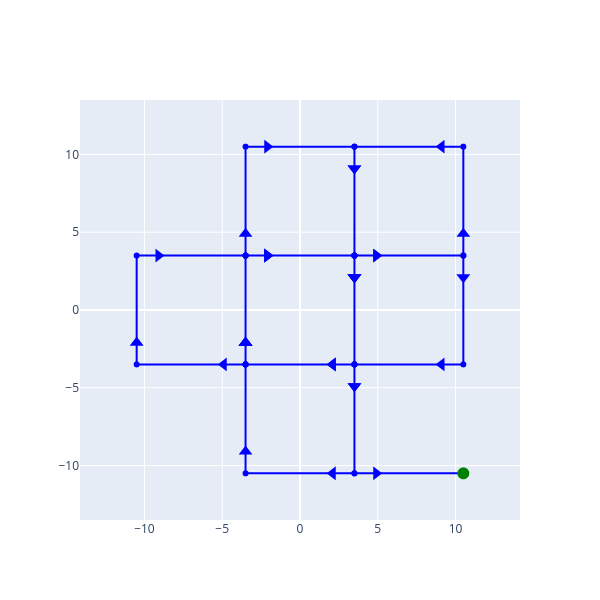}
        \caption{Corner segments}
    \end{subfigure}
    \begin{subfigure}[t]{0.19\textwidth}
        \includegraphics[width=\linewidth]{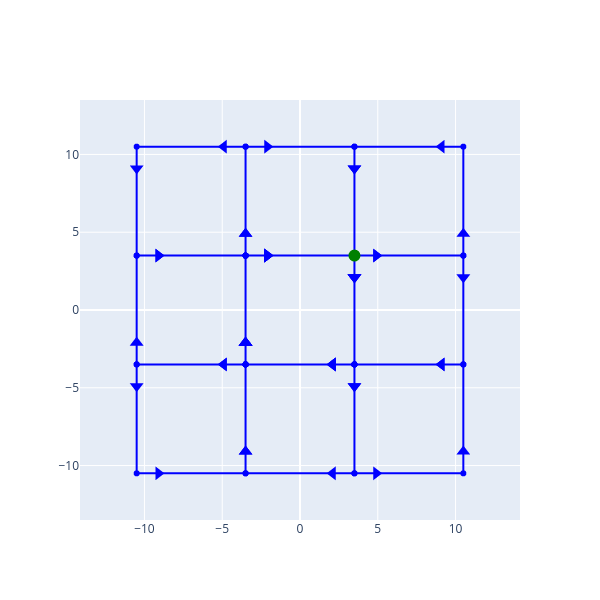}
        \caption{Full pattern}
    \end{subfigure}
    \caption{Progressive execution of the high-level path. The rover begins with a small loop and expands outward in overlapping square patterns to ensure complete coverage and frequent loop closures.}
    \label{fig:loop_path_frames}
\end{figure}


\subsection{Motion Planning}
\label{subsec:motion-planning}
The motion planner converts high-level waypoints into locally feasible trajectories that avoid collisions and minimize path length. 
Rather than maintaining an explicit costmap, we rely on direct geometric reasoning over detected rocks extracted from the segmentation masks. 

From the predicted semantic segmentation, we identify contiguous regions labeled as \textit{rock}. For each region, we compute its center and width in pixels, then convert this width to a physical radius using the known camera intrinsics and estimated depth:
\[
r_{\text{rock}} = \frac{w_{\text{px}} \cdot d}{2f}
\]
where $w_{\text{px}}$ is the width in pixels, $d$ is the depth to the center of the rock, and $f$ is the focal length in pixels. Rocks with radii below a minimum threshold are ignored to avoid overreacting to noise or small debris.

\begin{figure}[ht]
    \centering
    \begin{subfigure}[t]{0.25\textwidth}
        \includegraphics[width=\linewidth]{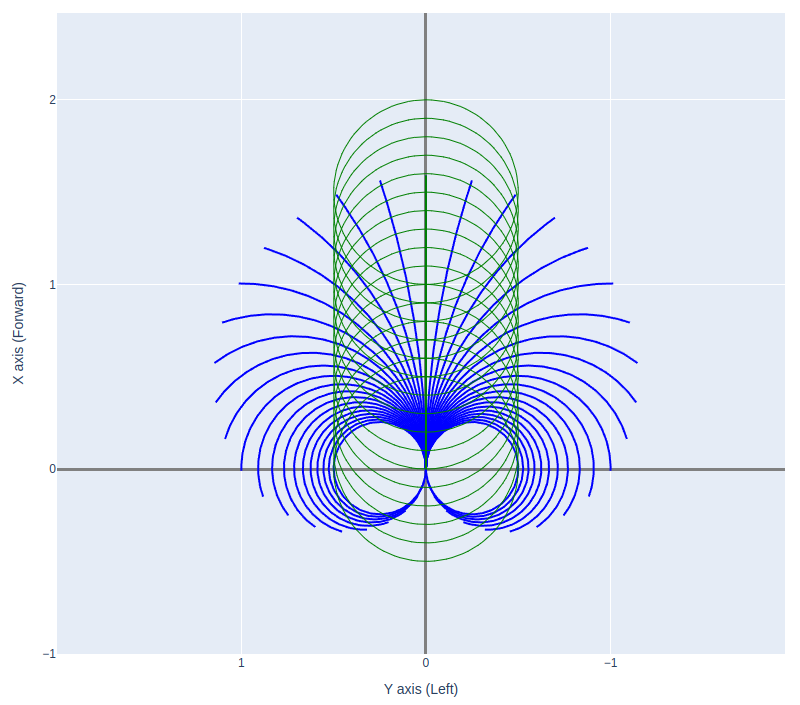}
        \caption{Candidate arcs in blue, with rover radius buffer shown in green.}
    \end{subfigure}
    \hfill
    \begin{subfigure}[t]{0.70\textwidth}
        \includegraphics[width=\linewidth]{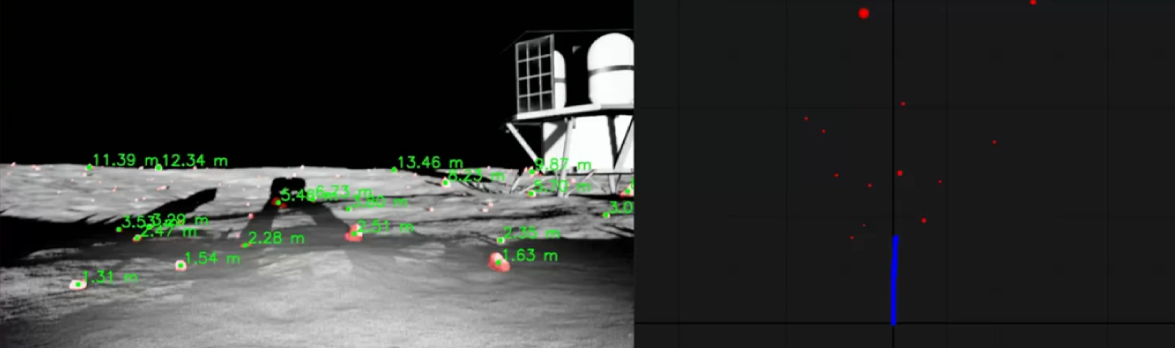}
        \caption{Example rock detection from semantic segmentation and corresponding rocks and planned arc placed in top-down 2D.}
    \end{subfigure}
    \caption{Example of detected rocks from stereo segmentation and depth estimates on the left, with planned arc and rocks shown in rover local frame on the right.}
    \label{fig:motion_planning}
\end{figure}

At each control cycle, we sample a discrete set of constant-curvature arcs, each representing a short forward motion with fixed angular velocity. For each candidate arc, we check whether the arc intersects with any detected rock (buffered by the rover radius), and discard any that do. Among the remaining arcs, we select the one whose endpoint is closest to the current goal waypoint:
\[
\text{argmin}_{\text{arc } i} \, \| \mathbf{p}_i^{\text{end}} - \mathbf{p}_{\text{goal}} \|_2
\]
This approach allows the planner to be fast, reactive, and interpretable while remaining robust to perceptual artifacts.
Figure~\ref{fig:motion_planning} shows the set of candidate arcs used as well as an example rock detection and planner arc.

\paragraph{Backup Maneuver.}
In cases where the motion planner fails to find a feasible path — due to unexpected obstacles, high slope, or visual drift — we implement a reactive backup maneuver to recover the rover’s motion. The trigger for this behavior is based on speed drop detection: if the rover’s actual speed falls below 25\% of the expected speed for more than 2–3 seconds, and no valid arc is found, the system initiates a timed reverse motion followed by a replan attempt.

This logic is carefully tuned to avoid false positives during normal slowdowns (e.g., when climbing uphill), and proved effective for recovering from collision during long-term autonomous runs. Fault recovery is also invoked when arc sampling fails entirely, acting as a last-resort behavior to regain mobility.

\section{Experiments and Results}
\label{sec:results}

We validate our full agent on the challenge scenario and benchmark its performance using the competition scoring.
The simulator provides two maps with different height geometry and lighting for development and validation, and a third undisclosed map is used for competition evaluation.
In addition, different rock presets are available and the agent's spawn location can be adjusted randomly.
All our code is available online at \texttt{\url{https://github.com/Stanford-NavLab/lunar_autonomy_challenge}}.


\subsection{Local Testing Results}

We evaluate the full system on Map 1 of the simulator, which includes access to ground truth maps and poses. This allows us to quantify both geometric and rock map quality as well as localization performance. Table~\ref{tab:1} shows our scores across five different rock distribution presets, which vary in rock density, placement, and size.

\begin{table}[htbp]
    \centering
    \caption{Performance across five rock distribution presets on Map 1. Our system consistently achieves strong localization and mapping scores across diverse terrain configurations.}
    \label{tab:1}
    \begin{tabular}{lcccc}
        \toprule
        \textbf{Preset} & \textbf{RMSE (m)} & \textbf{Geometric Map Score} & \textbf{Rock Map Score} & \textbf{Total Score} \\
        \midrule
        1 & 0.0434 & 269.6 & 153.6 & 823.3 \\
        2 & 0.0379 & 272.3 & 155.2 & 827.5 \\
        3 & 0.0605 & 200.8 & 146.2 & 746.9 \\
        4 & 0.0612 & 190.2 & 154.8 & 745.0 \\
        5 & 0.0510 & 224.7 & 150.6 & 775.3 \\
        \bottomrule
    \end{tabular}
\end{table}

To evaluate robustness to initial conditions, we repeat runs using the same rock preset (Preset 1) but with randomized initial positions. Results in Table~\ref{tab:2} demonstrate consistent localization and mapping performance across seeds.

\begin{table}[htbp]
    \centering
    \caption{Performance across different random seeds using the same rock preset. Our solution shows strong repeatability and robustness to initial rover placement.}
    \label{tab:2}
    \begin{tabular}{lcccc}
        \toprule
        \textbf{Preset} & \textbf{RMSE (m)} & \textbf{Geometric Map Score} & \textbf{Rock Map Score} & \textbf{Total Score} \\
        \midrule
        1 & 0.0628 & 195.7 & 150.4 & 746.1 \\
        1 & 0.0472 & 260.2 & 158.1 & 818.3 \\
        1 & 0.0671 & 221.0 & 144.7 & 765.7 \\
        1 & 0.0434 & 269.6 & 153.6 & 823.3 \\
        1 & 0.0510 & 262.8 & 151.8 & 814.6 \\
        \bottomrule
    \end{tabular}
\end{table}

\begin{figure}[H]
    \centering
    \begin{subfigure}[t]{0.49\textwidth}
        \includegraphics[width=\linewidth]{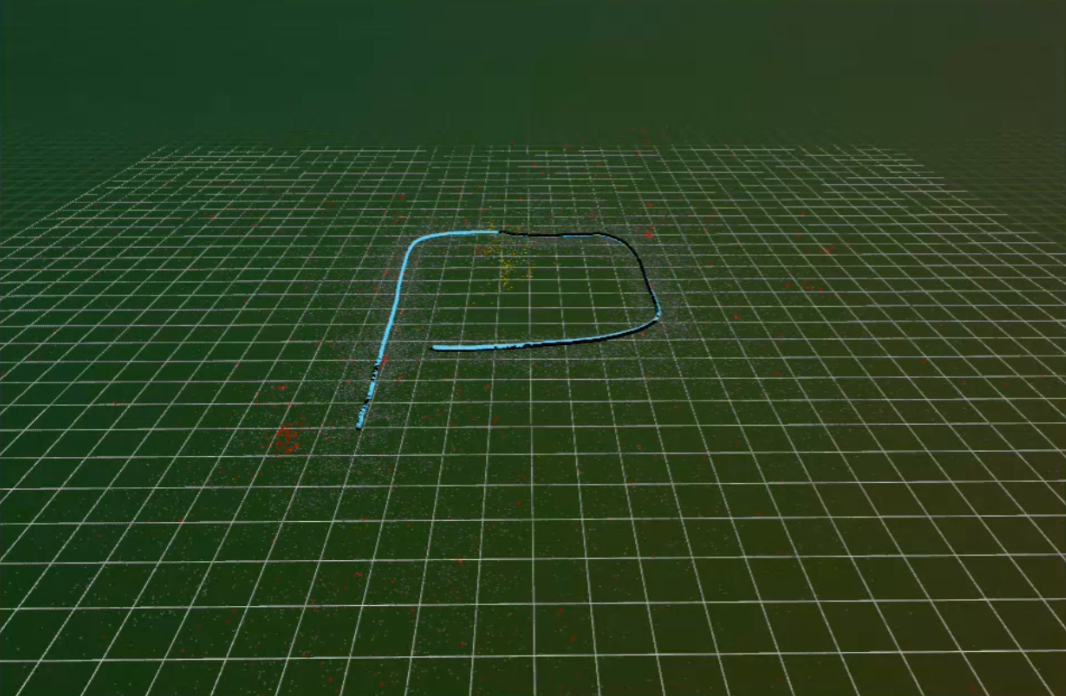}
    \end{subfigure}
    \begin{subfigure}[t]{0.49\textwidth}
        \includegraphics[width=\linewidth]{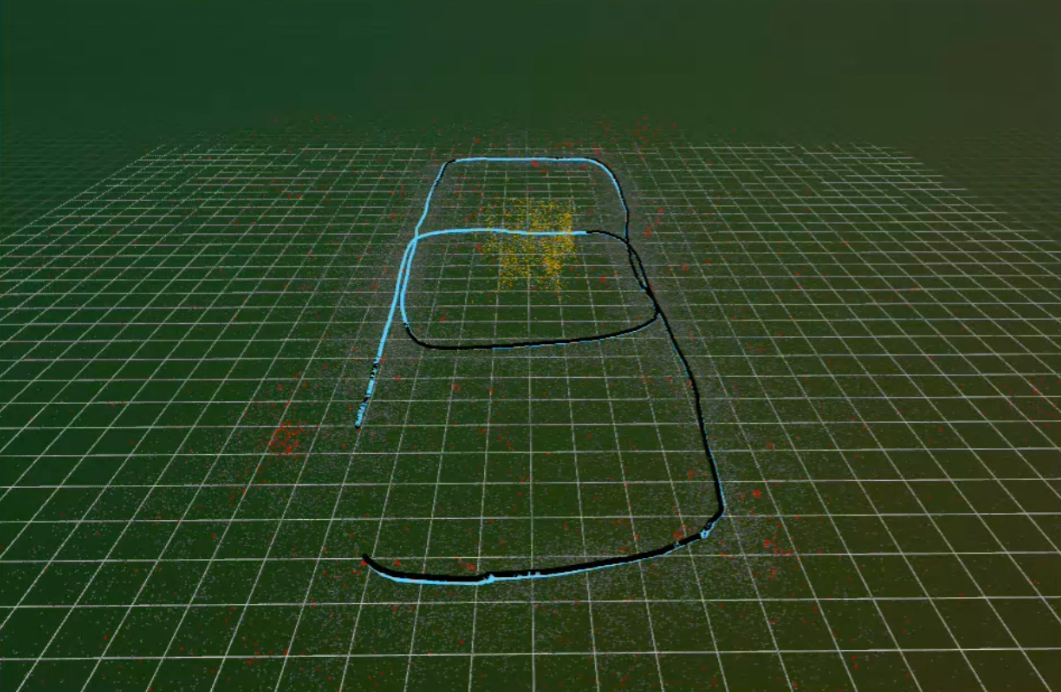}
    \end{subfigure} \\
    \begin{subfigure}[t]{0.49\textwidth}
        \includegraphics[width=\linewidth]{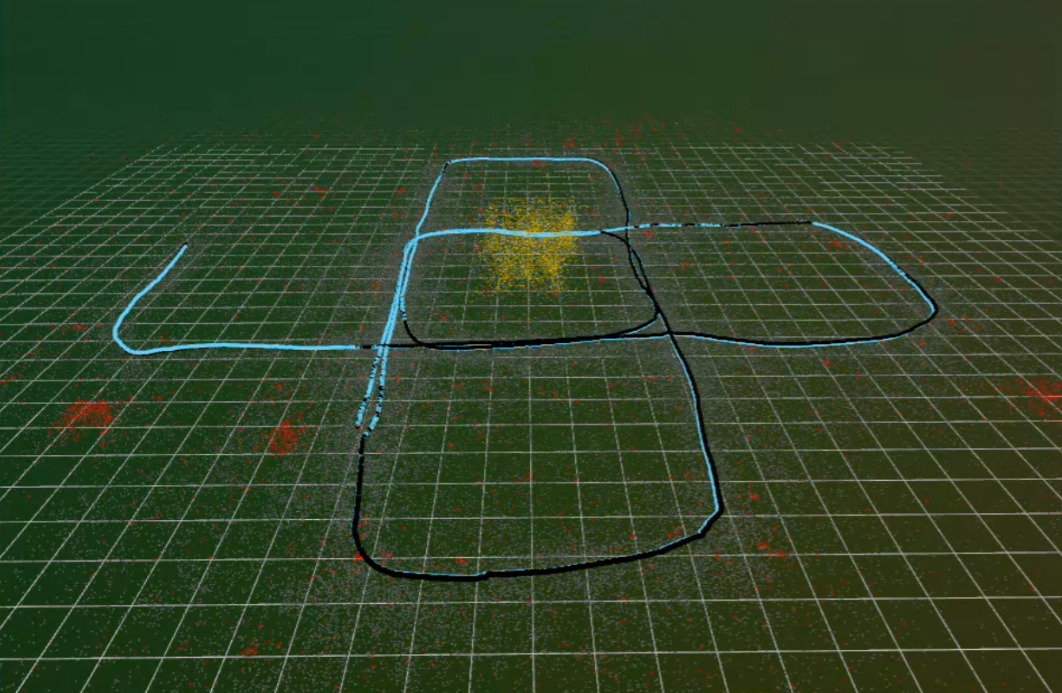}
    \end{subfigure}
    \begin{subfigure}[t]{0.49\textwidth}
        \includegraphics[width=\linewidth]{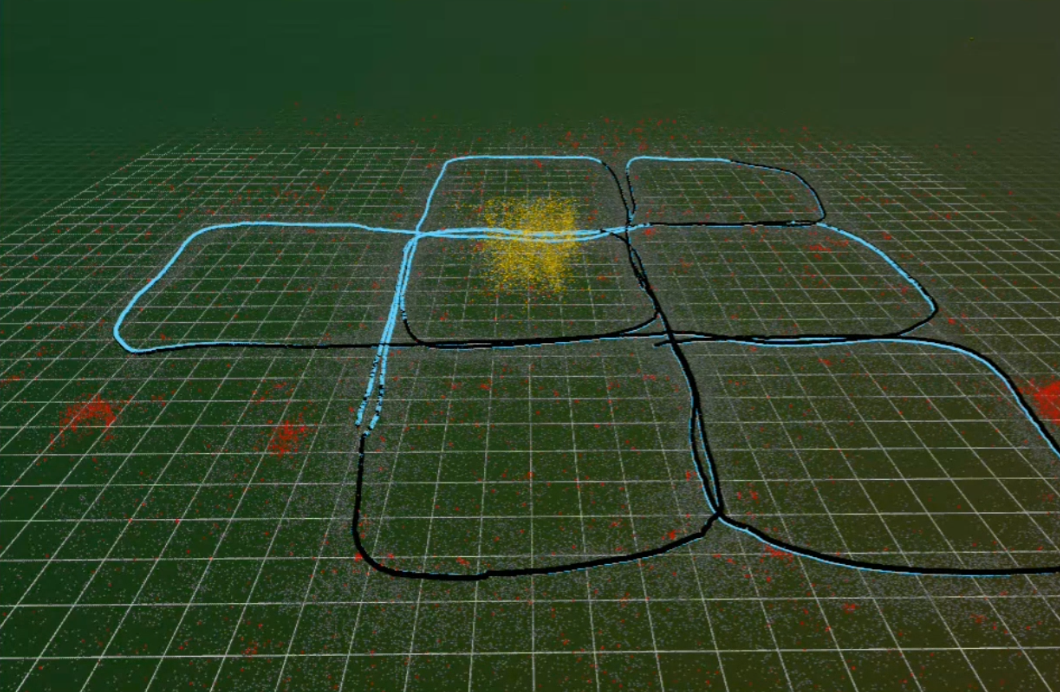}
    \end{subfigure}
    \caption{Progression of a sample run visualized using Rerun. The rover's trajectory (estimated in light blue, ground truth in black) is overlaid with the semantic point cloud. Ground points are shown in gray, rocks in red, and the lander in gold. The system maintains accurate localization and semantic consistency over the course of the mission.}
    \label{fig:rerun_mapping}
\end{figure}

\begin{figure}[ht]
    \centering
    \begin{subfigure}[t]{0.49\textwidth}
        \includegraphics[width=\linewidth]{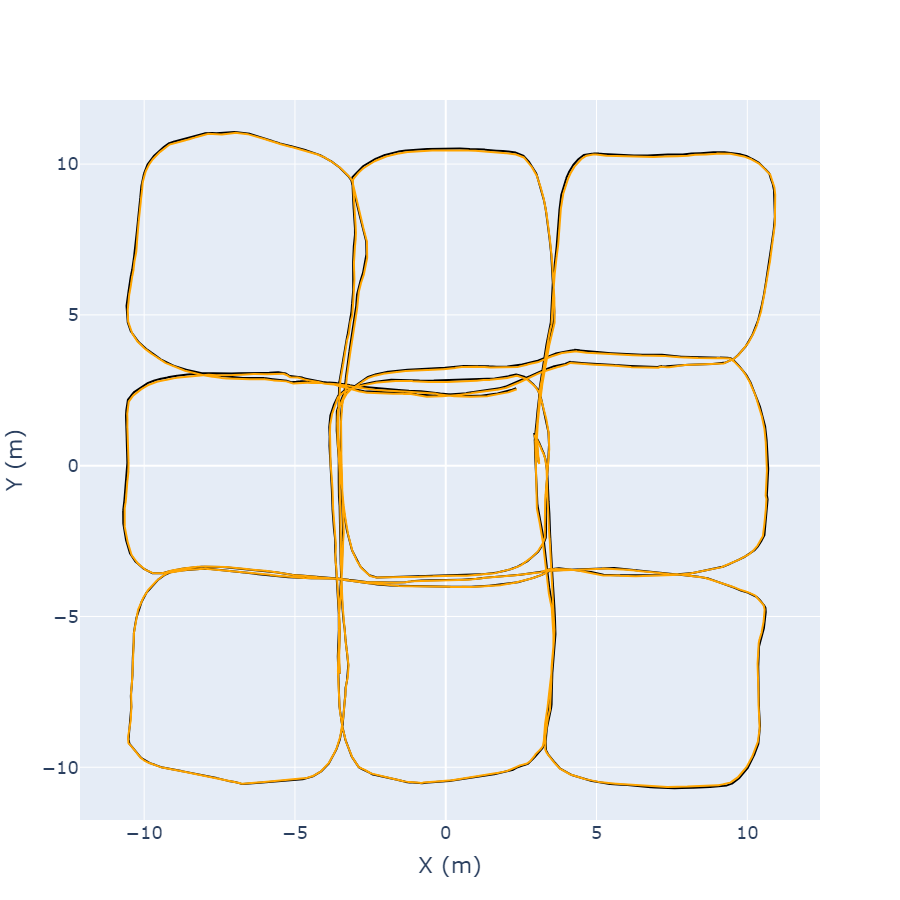}
    \end{subfigure}
    \hfill
    \begin{subfigure}[t]{0.49\textwidth}
        \includegraphics[width=\linewidth]{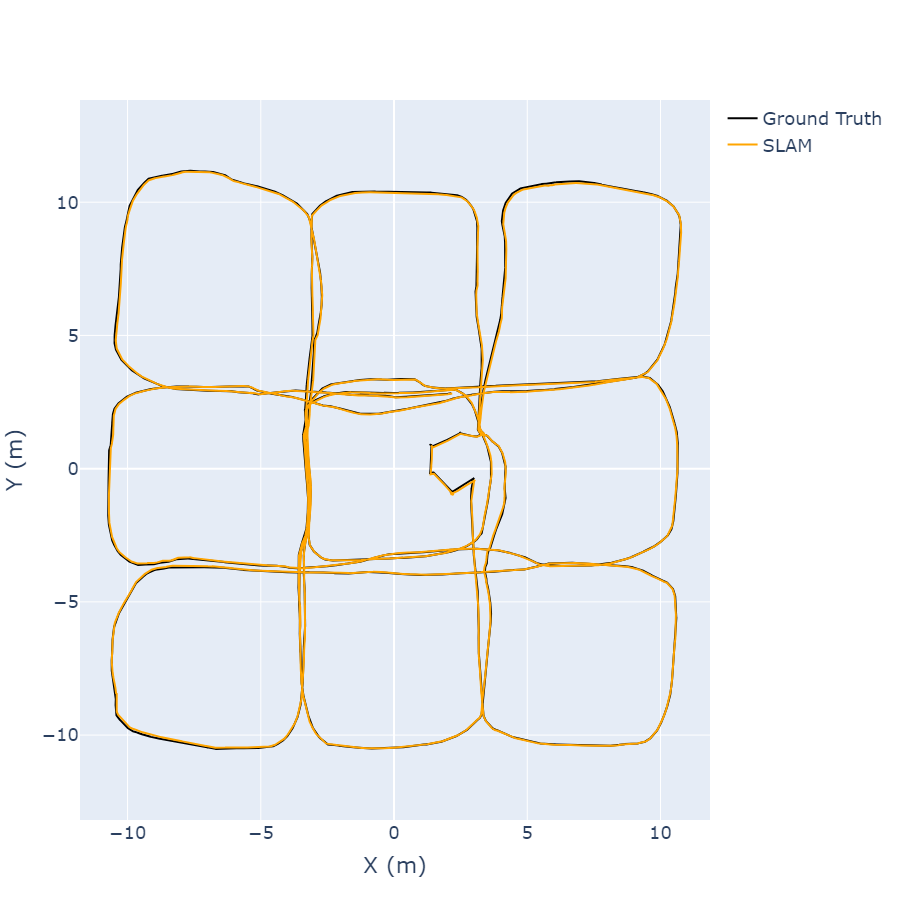}
    \end{subfigure}
    \caption{2D trajectories from two different runs, with ground truth trajectory shown in black and final estimated SLAM trajectory shown in orange. In the right plot, multiple backup maneuvers were triggered as the planner disengaged from local obstacles. Despite this, our SLAM maintains low localization error through the entire trajectory. }
    \label{fig:2d_traj_plots}
\end{figure}

\begin{figure}[H]
    \centering
    \begin{subfigure}[t]{0.49\textwidth}
        \includegraphics[width=\linewidth]{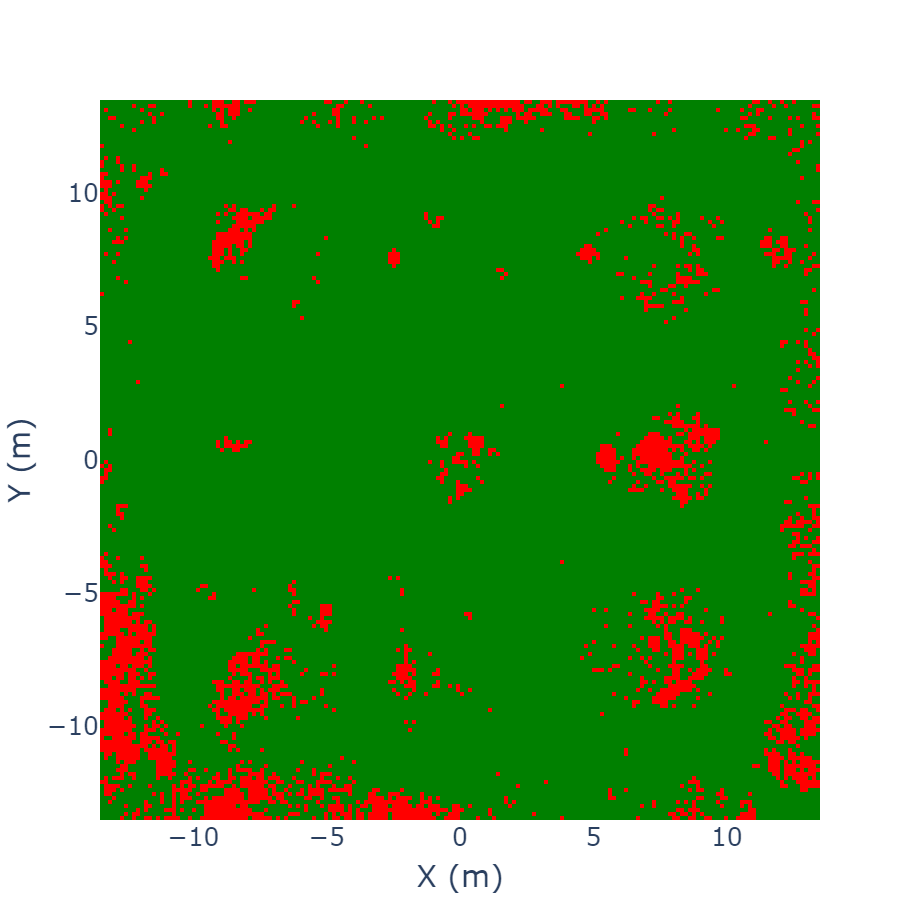}
        \caption{Geometric map visualization. Green cells are mapped within the 5 \si{cm} height error tolerance, and red cells exceed the error tolerance.}
    \end{subfigure}
    \hfill
    \begin{subfigure}[t]{0.49\textwidth}
        \includegraphics[width=\linewidth]{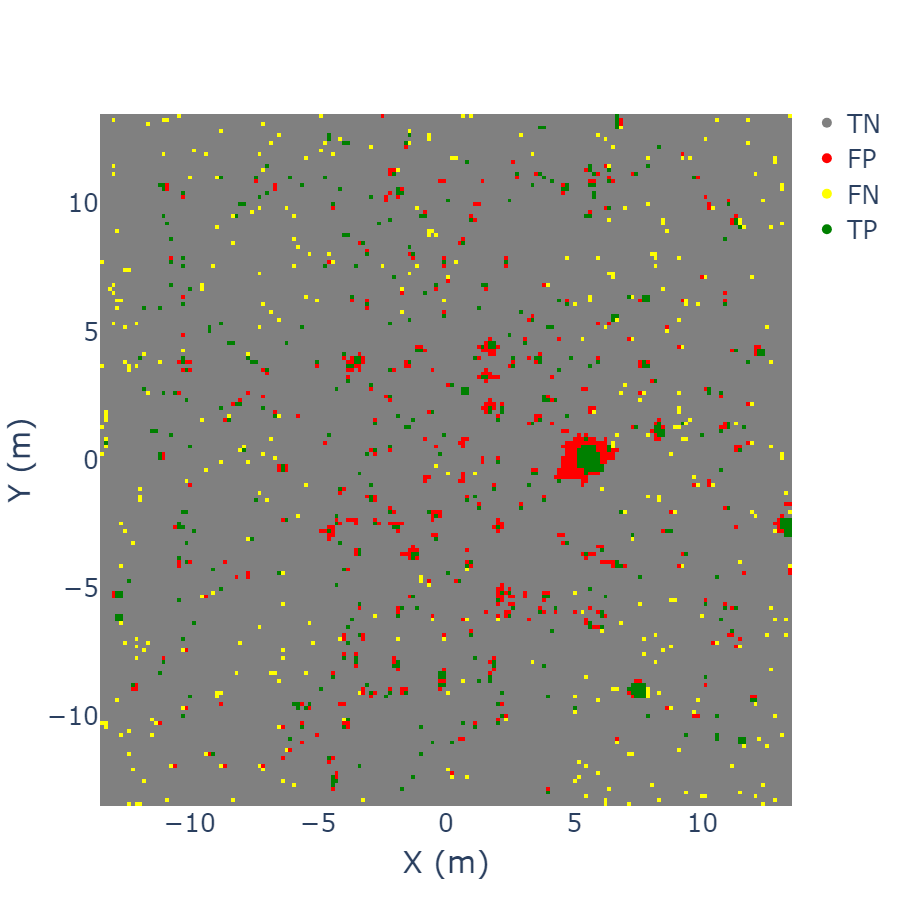}
        \caption{Rock map visualization. Cells are colored based on whether they were mapped as true positive (green), true negative (gray), false positive (red), and false negative (yellow).}
    \end{subfigure}
    \caption{Final geometric and rock maps from a representative run.}
    \label{fig:final_maps}
\end{figure}

\subsection{Segmentation}

To evaluate segmentation accuracy, we compare predicted masks against simulator-provided ground truth. We benchmark a variety of CNN and transformer-based models, as shown in Table~\ref{tab:semantic_segmentation_models}. We select U-Net++ for final deployment based on its strong tradeoff between accuracy and inference speed.


\begin{table*}[h]
    \centering
    \caption{Comparison of semantic segmentation performance across different models.}
    \label{tab:semantic_segmentation_models}
    \small
    \begin{tabular}{@{}lrrrrrr@{}}
        \toprule
        \multirow{2}{*}{\textbf{Model}} 
        & \multicolumn{1}{c}{\multirow{2}{*}{\textbf{Params}}} 
        & \multicolumn{1}{c}{\multirow{2}{*}{\textbf{FPS $\uparrow$}}} 
        & \multicolumn{4}{c}{\textbf{IoU [\%] $\uparrow$}} 
        \\
        \cmidrule{4-7}
        \multicolumn{3}{c}{} 
        & \textbf{Ground} 
        & \textbf{Rocks} 
        & \textbf{Sky} 
        & \textbf{Mean} 
        \\
        \midrule
        U-Net & 14.3M & 114.4 & 95.5 & 89.2 & \textbf{99.8} & 96.1 \\
        U-Net++ & 16.0M & 78.0 & 91.1 & \textbf{91.6} & \textbf{99.8} & 96.1 \\
        MA-Net & 21.7M & 71.7 & 70.9 & 90.1 & \textbf{99.8} & 91.5 \\
        Linknet & 11.7M & 104.5 & 96.4 & 91.1 & \textbf{99.8} & \textbf{96.7} \\
        FPN & 13.0M & 108.7 & \textbf{97.7} & 86.8 & 99.7 & 96.2 \\
        PSPNet & \textbf{0.9M} & \textbf{203.2} & 92.7 & 79.7 & 99.6 & 93.0 \\
        PAN & 11.4M & 92.8 & 94.6 & 84.3 & 99.7 & 95.0 \\
        DeepLabV3 & 15.9M & 131.5 & 96.7 & 86.6 & 99.7 & 96.1 \\
        DeepLabV3+ & 12.3M & 118.8 & 96.4 & 89.2 & \textbf{99.8} & 96.5 \\
        Segformer & 11.8M & 140.5 & 95.5 & 88.9 & \textbf{99.8} & 96.2 \\
        DPT & 41.6M & 79.6 & 94.1 & 87.4 & 99.7 & 95.2 \\
        \bottomrule
    \end{tabular}
\end{table*}

\subsection{Localization}

\begin{figure}[ht]
    \centering
    \includegraphics[width=0.9\linewidth]{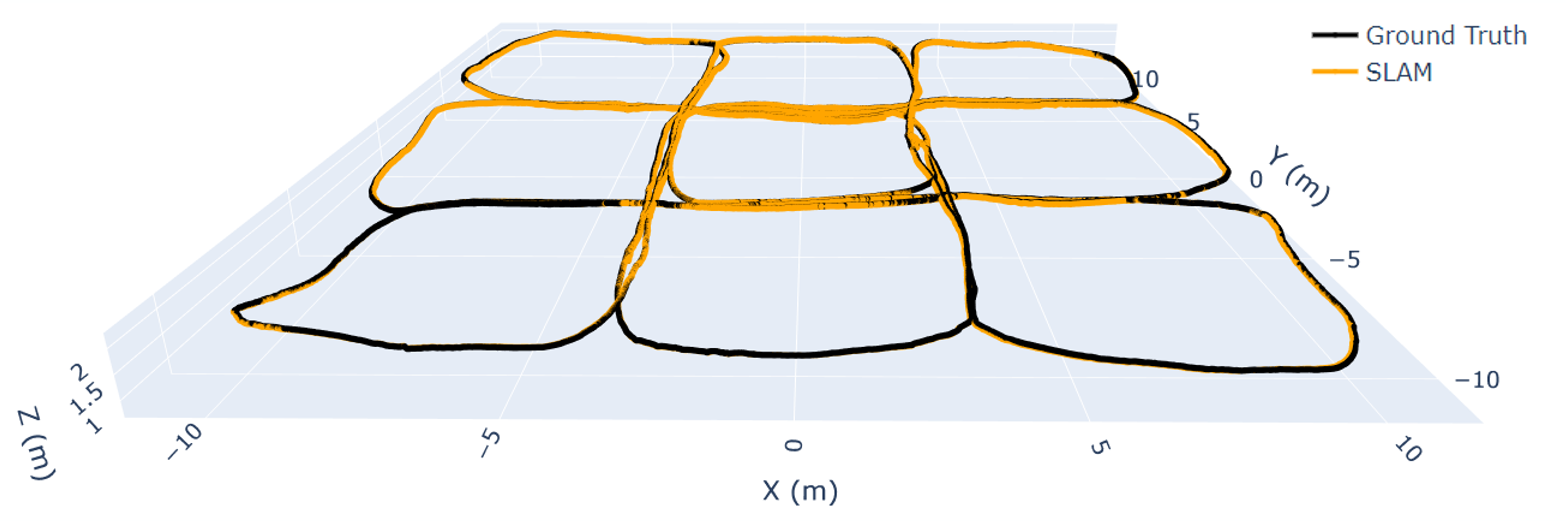}
    \caption{3D trajectory plot, with ground truth trajectory shown in black and final estimated SLAM trajectory shown in orange.} 
    \label{fig:traj_3d}
\end{figure}

\begin{figure}[H]
    \centering
    \includegraphics[width=\linewidth]{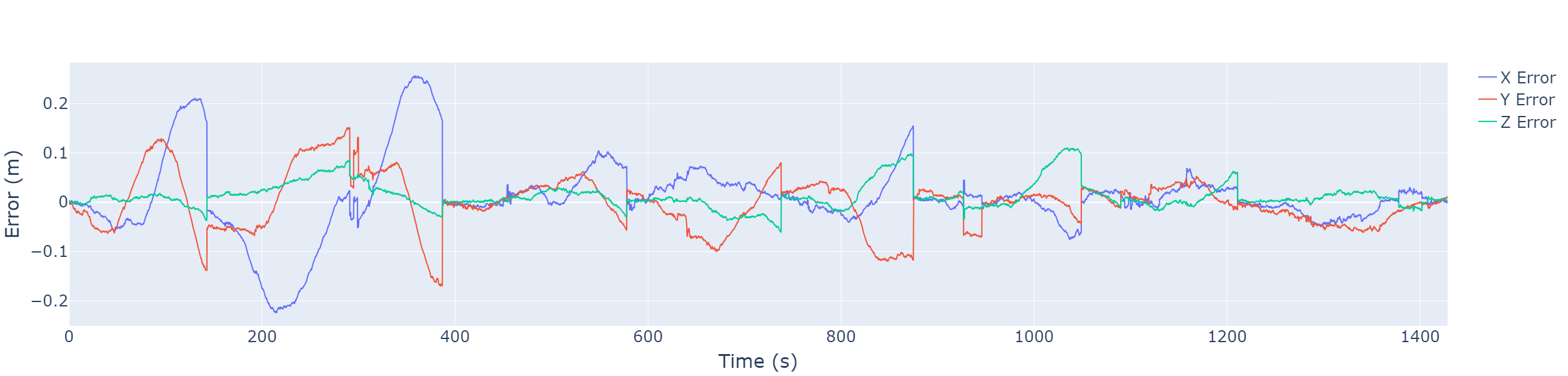}
    \caption{XYZ position errors over time (meters). Errors periodically drop due to loop closures 
    } 
    \label{fig:localization_error}
\end{figure}

Localization accuracy is measured using 3D position root mean square error (RMSE). Figure~\ref{fig:traj_3d} shows a 3D reconstruction of a representative trajectory, and Figure~\ref{fig:localization_error} plots the position error over time. We observe periodic error drops following successful loop closures.

As shown in Tables~\ref{tab:1} and~\ref{tab:2}, our system achieves consistent centimeter-level RMSE, which is critical to maintaining geometric mapping accuracy over long distances.



\subsection{Competition Results}

Our final submission to the Lunar Autonomy Challenge was evaluated on a hidden test map with unknown terrain geometry, lighting conditions, and rock configuration. Unlike development maps, this competition map was not accessible during testing and required the system to generalize across perceptual and planning components.

\begin{figure}[ht]
    \centering
    \includegraphics[width=0.8\linewidth]{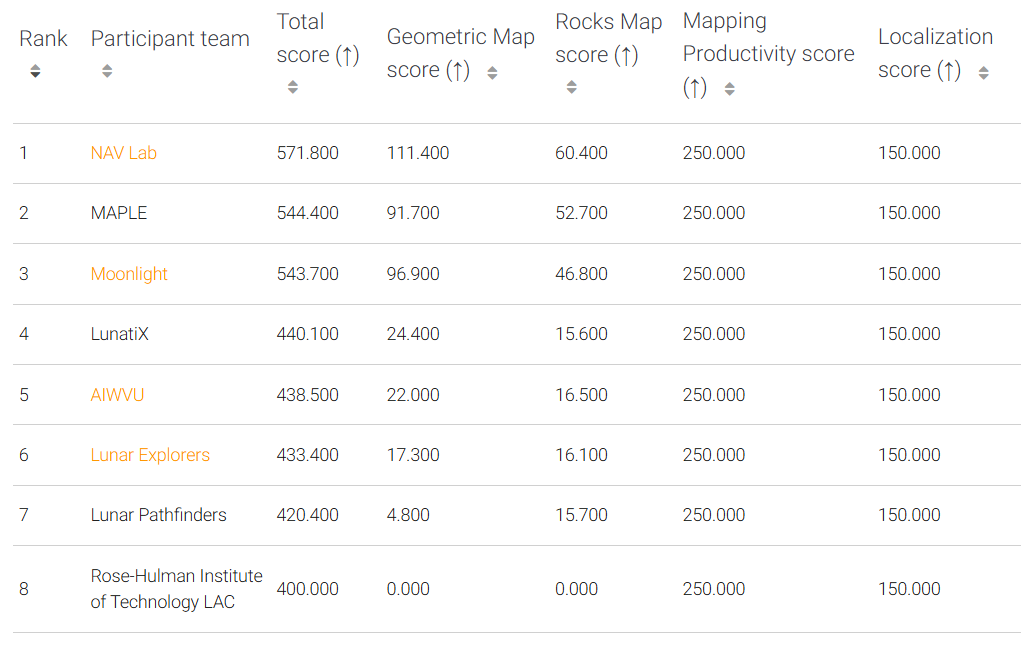}
    \caption{Final competition leaderboard. Our solution (NAV Lab) placed first with highest geometric and rock map scores.} 
    \label{fig:leaderboard}
\end{figure}

Figure~\ref{fig:leaderboard} shows the final competition leaderboard. Our solution, submitted as \textit{NAV Lab}, achieved the highest overall score out of all participating teams, placing first in both geometric mapping accuracy and rock detection performance. This demonstrates the robustness and generalization capability of our full-stack autonomy system, particularly under novel conditions.

Through experimentation, we found that the structured waypoint policy described in Section~\ref{subsec:path_design} performed well on development maps, but encountered failure modes on the competition map (for undetermined reasons, as detailed execution logs were not available). As a result, our highest scoring competition submission used an alternative outward spiral path centered around the lander. This design was meant to encourage loop closures between successive rings of the spiral while providing dense coverage of the inner mapping region.

Our success in the competition reflects not only the effectiveness of individual components such as SLAM and planning, but also the reliability of their integration in a modular and fault-tolerant pipeline.


\section{Conclusion}
\label{sec:conclusion}

We present a full-stack autonomous navigation system for lunar surface exploration, developed for the Lunar Autonomy Challenge. Our solution achieved 1st place overall in the competition, demonstrating robust and accurate performance across a range of simulated lunar environments. The system integrates semantic perception, stereo visual odometry, pose graph SLAM, and structured planning into a modular pipeline capable of centimeter-level localization and high-fidelity map generation.

Through extensive local testing and benchmark runs, we validate the repeatability and robustness of our approach under varying rock distributions and random initializations. Our design emphasizes modularity and reliability through a loosely coupled architecture. Although tightly coupled bundle adjustment methods can theoretically achieve higher accuracy by jointly optimizing structure and motion, we find that our factor-graph-based pose graph optimization strikes a favorable balance between performance and robustness. It allows us to exploit loop closures for drift correction without relying on fragile convergence behavior, and maintain pose graphs over tens of thousands of poses.

In future work, we plan to explore tighter integration between depth estimation and SLAM, adaptive path planning informed by real-time map confidence, and learned priors for dynamic hazard rejection. 
We are also interested in extending our semantic mapping to more expressive representations such as surface-constrained Gaussians to improve completeness and navigability in sparse or ambiguous terrain.

\section*{Acknowledgements}
We thank the organizers of the Lunar Autonomy Challenge—NASA, the Johns Hopkins University Applied Physics Laboratory (APL), Caterpillar Inc., and Embodied AI—for providing the simulation environment and support that enabled this work and the opportunity to participate in this challenge.
We also thank Daniel Neamati for insightful discussion and for reviewing the drafts of this paper. 

\clearpage

\printbibliography

\end{document}